INSTRUCTIONS:
The paper is in LaTeX source and uses the latex-acl and named style files.
(named.sty is a standard LaTeX style, latex-acl.sty can be retrieved from the
 macro collection of the cmp-lg server).
The bibliography file is appended following the latex file.

- Seperate the two files, saving the first as "paper.tex" and the second as
  "paper.bbl"
- Run latex on "paper.tex" twice (to get the cross-references right)
  (note there is no need to run bibtex).
- Run dvips or xdvi (or whatever software you use to view the .dvi file)

---------- latex file "paper.tex" starts here --------------------------------
\documentstyle[latex-acl,named]{article}

\begin{document}

\title{AN INTEGRATED HEURISTIC SCHEME \\
       FOR PARTIAL PARSE EVALUATION}

\author{Alon Lavie \\
        School of Computer Science\\
        Carnegie Mellon University\\
        5000 Forbes Ave., Pittsburgh, PA 15213\\
        {\tt email : lavie@cs.cmu.edu}}

\maketitle

\begin{abstract}
GLR* is a recently developed robust version of the Generalized LR Parser
\cite{tomita1}, that can parse almost {\em any} input sentence by ignoring
unrecognizable parts of the sentence. On a given input sentence, the parser
returns a collection of parses that correspond to maximal, or close to maximal,
parsable subsets of the original input. This paper describes recent work on
developing an integrated heuristic scheme for selecting the parse that is
deemed ``best'' from such a collection. We describe the heuristic measures
used and their combination scheme. Preliminary results from experiments
conducted on parsing speech recognized spontaneous speech are also reported.
\end{abstract}

\section{The GLR* Parser}

\subsection{The GLR Parsing Algorithm}

The Generalized LR Parser, developed by Tomita \cite{tomita1}, extended the
original LR parsing algorithm to the case of non-LR languages, where the
parsing tables contain entries with multiple parsing actions.
Tomita's algorithm uses a Graph Structured Stack (GSS) in order to
efficiently pursue in parallel the different parsing options that arise
as a result of the multiple entries in the parsing tables. A second data
structure uses pointers to keep track of all possible parse trees throughout
the parsing of the input, while sharing common subtrees of these different
parses. A process of local ambiguity packing allows the parser to pack
sub-parses that are rooted in the same non-terminal into a single structure
that represents them all.

The GLR parser is the syntactic engine of the Universal Parser Architecture
developed at CMU \cite{tomita3}. The architecture supports grammatical
specification in an LFG framework, that consists of context-free grammar rules
augmented with feature bundles that are associated with the non-terminals of
the rules. Feature structure computation is, for the most part, specified and
implemented via unification operations. This allows the grammar to constrain
the applicability of context-free rules. The result of parsing an input
sentence consists of both a parse tree and the computed feature structure
associated with the non-terminal at the root of the tree.

\subsection{The GLR* Parser}

GLR* is a recently developed robust version of the Generalized LR Parser, that
allows the skipping of unrecognizable parts of the input sentence
\cite{lavie1}. It is designed to enhance the parsability of domains such as
spontaneous speech, where the input is likely to contain deviations from the
grammar, due to either extra-grammaticalities or limited grammar coverage.
In cases where the complete input sentence is not covered by the grammar,
the parser attempts to find a maximal subset of the input that is parsable.
In many cases, such a parse can serve as a good approximation to the true parse
of the sentence.

The parser accommodates the skipping of words of the input string by allowing
shift operations to be performed from inactive state nodes in the Graph
Structured Stack (GSS). Shifting an input symbol from an inactive state is
equivalent to skipping the words of the input that were encountered after the
parser reached the inactive state and prior to the current word that is being
shifted. Since the parser is LR(0), previous reduce operations remain
valid even when words further along in the input are skipped. Information
about skipped words is maintained in the symbol nodes that represent parse
sub-trees.

To guarantee runtime feasibility, the GLR* parser is coupled with a ``beam''
search heuristic, that dynamically restricts the skipping capability of the
parser, so as to focus on parses of maximal and close to maximal substrings
of the input. The efficiency of the parser is also increased by an enhanced
process of local ambiguity packing and pruning. Locally ambiguous symbol nodes
are compared in terms of the words skipped within them. In cases where one
phrase has more skipped words than the other, the phrase with more skipped
words is discarded in favor of the more complete parsed phrase. This operation
significantly reduces the number of parses being pursued by the parser.

\section{The Parse Evaluation Heuristics}

At the end of the process of parsing a sentence, the GLR* parser returns with
a set of possible parses, each corresponding to some grammatical subset of
words of the input sentence. Due to the beam search heuristic and the
ambiguity packing scheme, this set of parses is limited to maximal or close to
maximal grammatical subsets. The principle goal is then to find the maximal
parsable subset of the input string (and its parse). However, in many cases
there are several distinct maximal parses, each consisting of a different
subset of words of the original sentence. Furthermore, our experience has
shown that in many cases, ignoring an additional one or two input words may
result in a parse that is syntactically and/or semantically more coherent. We
have thus developed an evaluation heuristic that combines several different
measures, in order to select the parse that is deemed overall ``best''.

Our heuristic uses a set of features by which each of the parse candidates can
be evaluated and compared. We use features of both the candidate parse and the
ignored parts of the original input sentence. The features are designed to be
general and, for the most part, grammar and domain independent. For each parse,
the heuristic computes a penalty score for each of the features. The penalties
of the different features are then combined into a single score using a linear
combination. The weights used in this scheme are adjustable, and can be
optimized for a particular domain and/or grammar. The parser then selects the
parse ranked best (i.e. the parse of lowest overall score).~\footnote{The
system can display the $n$ best parses found, where the parameter $n$ is
controlled by the user at runtime. By default, we set $n$ to one, and the
parse with the lowest score is displayed.}

\subsection{The Parse Evaluation Features}

So far, we have experimented with the following set of evaluation features:
\begin{enumerate}
\item The number and position of skipped words
\item The number of substituted words
\item The fragmentation of the parse analysis
\item The statistical score of the disambiguated parse tree
\end{enumerate}

The penalty scheme for skipped words is designed to prefer parses that
correspond to fewer skipped words. It assigns a penalty in the range of
$(0.95 - 1.05)$ for each word of the original sentence that was skipped.
The scheme is such that words that are skipped later in the sentence receive
the slightly higher penalty. This preference was designed to handle the
phenomena of false starts, which is common in spontaneous speech.

The GLR* parser has a capability for handling common word substitutions when
the parser's input string is the output of a speech recognition system. When
the input contains a pre-determined commonly substituted word, the parser
attempts to continue with both the original input word and a specified
``correct'' word. The number of substituted words is used as an evaluation
feature, so as to prefer an analysis with fewer substituted words.

The grammars we have been working with allow a single input sentence to be
analyzed as several grammatical ``sentences'' or fragments. Our experiments
have indicated that, in most cases, a less fragmented analysis is more
desirable. We therefore use the sum of the number of fragments in the
analysis as an additional feature.

We have recently augmented the parser with a statistical disambiguation module.
We use a framework similar to the one proposed by Briscoe and Carroll
\cite{briscoe1}, in which the shift and reduce actions of the LR parsing
tables are directly augmented with probabilities. Training of the probabilities
is performed on a set of disambiguated parses. The probabilities of the parse
actions induce statistical scores on alternative parse trees, which are used
for disambiguation. However, additionally, we use the statistical
score of the disambiguated parse as an additional evaluation feature
{\em across} parses. The statistical score value is first converted into a
confidence measure, such that more ``common'' parse trees receive a lower
penalty score. This is done using the following formula:

\[ penalty = (0.1 * (- log_{10}(pscore))) \]

The penalty scores of the features are then combined by a linear
combination. The weights assigned to the features determine the way they
interact. In our experiments so far, we have fined tuned these weights
manually, so as to try and optimize the results on a training set of data.
However, we plan on investigating the possibility of using some known
optimization techniques for this task.

\subsection{The Parse Quality Heuristic}

The utility of a parser such as GLR* obviously depends on the semantic
coherency of the parse results that it returns. Since the parser is designed
to succeed in parsing almost any input, parsing success by itself can no
longer provide a likely guarantee of such coherency. Although we believe this
task would ultimately be better handled by a domain dependent semantic analyzer
that would follow the parser, we have attempted to partially handle this
problem using a simple filtering scheme.

\begin{table*}[t]
\centering
\begin{tabular}{||l|c|c|c|c||c|c|c|c||} \hline
\mbox{} & \multicolumn{2}{|c|}{Unparsable} &
\multicolumn{2}{|c||}{Parsable} & \multicolumn{2}{|c|}{Good/Close} &
\multicolumn{2}{|c||}{Bad} \\
\mbox{} & \multicolumn{2}{|c|}{\mbox{}} &
\multicolumn{2}{|c||}{\mbox{}} & \multicolumn{2}{|c|}{Parses} &
\multicolumn{2}{|c||}{Parses} \\ \hline
\mbox{} & number & percent & number & percent & number & percent &
number & percent \\ \hline
GLR  & 58 & 48.3\% & 62 & 51.7\% & 60 & 50.0\% & 2 & 1.7\% \\ \hline
GLR* (1) & 5 & 4.2\% & 115 & 95.8\% & 84 & 70.0\% & 31 & 25.8\% \\ \hline
GLR* (2) & 5 & 4.2\% & 115 & 95.8\% & 90 & 75.0\% & 25 & 20.8\% \\ \hline
\end{tabular}
\caption{Performance Results of the GLR* Parser}
(1) = simple heuristic, (2) = full heuristics
\label{perform-tab1}
\end{table*}

The filtering scheme's task is to classify the parse chosen as best by the
parser into one of two categories: ``good'' or ``bad''. Our heuristic takes
into account both the actual value of the parse's combined penalty score and a
measure relative to the length of the input sentence. Similar to the penalty
score scheme, the precise thresholds are currently fine tuned to try and
optimize the classification results on a training set of data.

\section{Parsing of Spontaneous Speech Using GLR*}

We have recently conducted some new experiments to test the utility of the GLR*
parser and our parse evaluation heuristics when parsing speech recognized
spontaneous speech in the ATIS domain. We modified an
existing partial coverage syntactic grammar into a grammar for the ATIS domain,
using a development set of some 300 sentences. The resulting grammar has 458
rules, which translate into a parsing table of almost 700 states.

A list of common appearing substitutions was constructed from the development
set. The correct parses of 250 grammatical sentences were used to train the
parse table statistics that are used for disambiguation and parse evaluation.
After some experimentation, the evaluation feature weights were set in the
following way. As previously described, the penalty for a skipped word ranges
between 0.95 and 1.05, depending on the word's position in the sentence. The
penalty for a substituted word was set to 0.9, so that substituting a word
would be preferable to skipping the word. The fragmentation feature was
given a weight of 1.1, to prefer skipping a word if it reduces the
fragmentation count by at least one. The three penalties are then summed,
together with the converted statistical score of the parse.

We then used a set of 120 new sentences as a test set. Our goal was three-fold.
First, we wanted to compare the parsing capability of the GLR* parser with
that of the original GLR parser. Second, we wished to test the effectiveness
of our evaluation heuristics in selecting the best parse. Third, we wanted to
evaluate the ability of the parse quality heuristic to correctly classify
GLR* parses as ``good'' or ``bad''. We ran the parser three times on the test
set. The first run was with skipping disabled. This is equivalent to running
the original GLR parser. The second run was conducted with skipping enabled
and full heuristics. The third run was conducted with skipping enabled, and
with a simple heuristic that prefers parses based only on the number of words
skipped. In all three runs, the single selected parse result for each sentence
was manually evaluated to determine if the parser returned with a ``correct''
parse.

The results of the experiment can be seen in Table~\ref{perform-tab1}.
The results indicate that using the GLR* parser results in a significant
improvement in performance. When using the full heuristics, the percentage of
sentences, for which the parser returned a parse that matched or almost
matched the ``correct'' parse increased from 50\% to 75\%. As a result of its
skipping capabilities, GLR* succeeds to parse 58 sentences (48\%) that were
not parsable by the original GLR parser. Fully 96\% of the test sentences (all
but 5) are parsable by GLR*. However, a significant portion of these sentences
(23 out of the 58) return with bad parses, due to the skipping of essential
words of the input. We looked at the effectiveness of our parse quality
heuristic in identifying such bad parses. The heuristic is successful in
labeling 21 of the 25 bad parses as ``bad''. 67 of the 90 good/close parses
are labeled as ``good'' by the heuristic. Thus, although somewhat overly harsh,
the heuristic is quite effective in identifying bad parses.

Our results indicate that our full integrated heuristic scheme for selecting
the best parse out-performs the simple heuristic, that considers only the
number of words skipped. With the simple heuristic, good/close parses were
returned in 24 out of the 53 sentences that involved some degree of skipping.
With our integrated heuristic scheme, good/close parses were returned in 30
sentences (6 additional sentences). Further analysis showed that only 2
sentences had parses that were better than those selected by our integrated
parse evaluation heuristic.

\bibliographystyle{named}

\bibliography{biblist}

\end{document}